\documentclass[sigconf]{acmart}

\copyrightyear{2026}
\acmYear{2026}
\setcopyright{cc}
\setcctype{by}
\acmConference[WWW '26] {Proceedings of the ACM Web Conference 2026}{April 13--17, 2026}{Dubai, United Arab Emirates.}
\acmBooktitle{Proceedings of the ACM Web Conference 2026 (WWW '26), April 13--17, 2026, Dubai, United Arab Emirates}
\acmDOI{10.1145/3774904.3792158}
\acmISBN{979-8-4007-2307-0/2026/04}

\author{Qianchi Zhang}
\authornote{This work was supported by Beijing Advanced Innovation Center for Future Blockchain and Privacy Computing.}
\affiliation{%
  \institution{School of Artificial Intelligence, Beihang University}
  \city{Beijing}
  \country{China}
}
\email{zhangqianchi@buaa.edu.cn}

\author{Hainan Zhang}
\authornote{Corresponding author.}
\affiliation{%
  \institution{School of Artificial Intelligence, Beihang University}
  \city{Beijing}
  \country{China}}
\email{zhanghainan@buaa.edu.cn}

\author{Liang Pang}
\affiliation{
  \institution{Institute of Computing Technology, Chinese Academy of Sciences}
  \city{Beijing}
  \country{China}
}
\email{pangliang@ict.ac.cn}

\author{Yongxin Tong}
\affiliation{%
  \institution{School of Computer Science and Engineering, Beihang University}
  \city{Beijing}
  \country{China}}
  \email{yxtong@buaa.edu.cn}

\author{Hongwei Zheng}
\affiliation{
 \institution{Beijing Academy of Blockchain and Edge Computing}
 \city{Beijing}
 \country{China}}
 \email{zhenghongwei2024@163.com}

\author{Zhiming Zheng}
\affiliation{%
  \institution{School of Artificial Intelligence, Beihang University}
  \city{Beijing}
  \country{China}}
  \email{zhengzhiming0130@163.com}

\usepackage{xcolor}
\usepackage{url}
\usepackage{arydshln}
\usepackage{xspace}
\usepackage{multirow} 
\usepackage{float}
\usepackage{adjustbox}
\usepackage{tcolorbox}  
\usepackage[table]{xcolor}
\usepackage{arydshln}
\usepackage{graphicx}
\usepackage{subcaption}  
\usepackage{rotating}
\usepackage{balance}
\begin{document}

\title{Less is More: Compact Clue Selection for Efficient Retrieval-Augmented Generation Reasoning}



\begin{abstract}
Current RAG retrievers are designed primarily for human readers, emphasizing complete, readable, and coherent paragraphs. However, Large Language Models (LLMs) benefit more from precise, compact, and well-structured input, which enhances reasoning quality and efficiency. Existing methods rely on reranking or summarization to identify key sentences, but may introduce semantic breaks and unfaithfulness. Thus, efficiently extracting and organizing answer-relevant clues from large-scale documents while reducing LLM reasoning costs remains challenging in RAG systems. Inspired by Occam's razor, we frame LLM-centric retrieval as MinMax optimization: maximizing the extraction of potential clues and reranking them for well-organization, while minimizing reasoning costs by truncating to the smallest sufficient set of clues. In this paper, we propose CompSelect, a compact clue selection mechanism for LLM-centric RAG, consisting of a clue extractor, a reranker, and a truncator. (1) The clue extractor first uses answer-containing sentences as fine-tuning targets, aiming to extract \textbf{sufficient} potential clues; (2) The reranker is trained to prioritize \textbf{effective} clues based on real LLM feedback; (3) The truncator uses the truncated text containing the minimum sufficient clues for answering the question as fine-tuning targets, thereby enabling \textbf{efficient} RAG reasoning. Experiments on three QA datasets demonstrate that CompSelect improves performance while reducing both total and online latency compared to a range of baseline methods. Further analysis also confirms its robustness to unreliable retrieval and generalization across different scenarios.
\end{abstract}


\maketitle

\section{Introduction}

\begin{figure}[!t]
\centering
  \includegraphics[width=1.0\columnwidth]{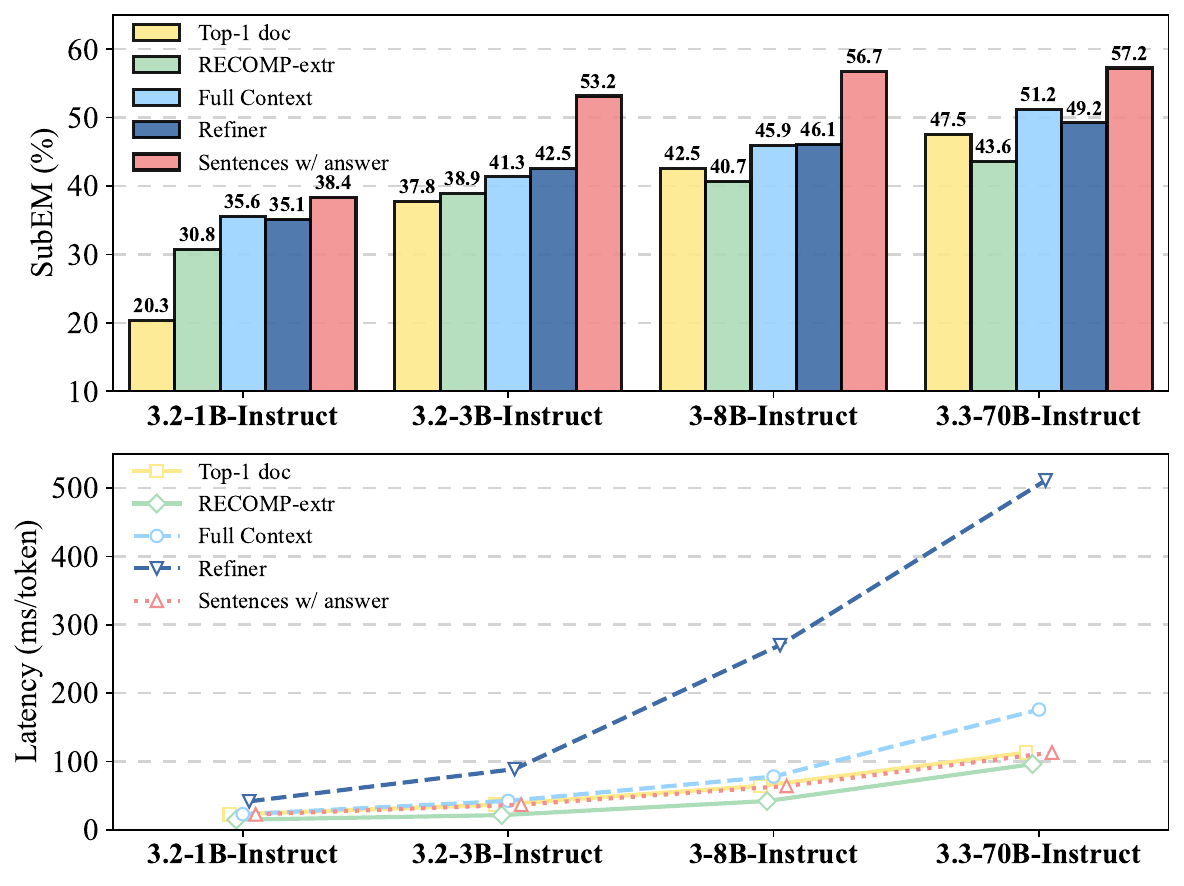}
\caption{SubEM performance and latency on the NQ test set across different LLaMA3 model sizes, including the reranking method RECOMP-extr and the abstractive method Refiner.}
  \label{fig:figure1}
\end{figure}

Retrieval-Augmented Generation (RAG)~\cite{lewis2020retrieval,gao2023retrieval,trustworth} has emerged as a promising paradigm to enhance Large Language Models (LLMs) with external knowledge for various tasks~\cite{chen2022gere,huang2023learning,chen2025privacy,wang2025learning}. However, the retrievers currently used in RAG are designed primarily for human readers, favoring long and coherent passages to enhance readability. This paradigm is less effective when the downstream consumer is an LLM~\cite{sauchuk2022role}, which typically performs better when provided with precise, compact, and well-structured inputs. As shown in Figure~\ref{fig:figure1}, we compare the Top-1 document, the full set of retrieved documents (approximately 5), and the answer-containing sentences (Oracle Clues) on the NQ dataset across different LLaMA3 model sizes. The results indicate that, regardless of whether it is a 1B model or the larger 70B model, structuring retrieved documents into a concise and efficient form (as Oracle Clues) consistently improves reasoning performance and reduces system latency. These findings highlight the need for compact clue selection mechanisms tailored to LLM reasoning in RAG systems.

Prior work has explored techniques such as reranking~\cite{xu2024recomp,ke-etal-2024-bridging,qin-etal-2024-large} and summarization~\cite{exit,zhu2024information} to help RAG identify key sentences. 
The former reranks the sentences from retrieved documents based on metrics such as answer contribution or user preference~\cite{mao2016does,zhu2023large}. 
The latter identifies query-relevant sentences either through extractive selection or abstractive generation.
Each approach has limitations: sentence-level reranking may disrupt the semantic structure of documents; extractive methods are limited in integrating dispersed information and capturing global context; and abstractive methods increase inference costs and may alter the original documents, thereby introducing issues of unfaithfulness.
As shown in Figure~\ref{fig:figure1}, the reranking method RECOMP-extr~\cite{xu2024recomp} primarily reduces reasoning latency but sacrifices QA performance, whereas the abstractive method Refiner~\cite{refiner} improves QA performance but increases system latency compared to baseline models. Therefore, efficiently extracting and organizing answer-relevant clues from large-scale documents while reducing LLM reasoning overhead remains an open challenge for RAG~\cite{deepagent,su2025clue}.

Inspired by Occam’s razor~\cite{ockham1,occam2}, we frame LLM-centric retrieval as a MinMax optimization problem. We first leverage contextual information to identify potential clues, forming the maximal subset capable of answering the query. Next, we rerank these clues based on their completeness and relevance to the query, moving the most effective ones at the forefront. Finally, we truncate the set to retain only the essential clues, thereby minimizing the context required for RAG. 

In this paper, we propose \textbf{CompSelect}, a compact clue selection mechanism for LLM-centric RAG, consisting of three modules: a clue extractor, a reranker, and a truncator. The extractor and reranker work together to provide sufficient and effective reasoning clues, while the truncator reduces reasoning cost. 
Specifically, (1) the clue extractor is fine-tuned using all sentences containing the answer and their semantically similar sentences based on K-Nearest Neighbors (KNN) clustering~\cite{guo2003knn,peterson2009k}, since RAG often requires richer contextual information for multi-hop reasoning. This ensures the extractor can capture \textbf{sufficient} potential clues. (2) The reranker is trained with real feedback from the LLM generator, labeling clues that lead to correct answers as positive samples and others as negative, enabling it to prioritize \textbf{effective} clues. 
(3) The truncator is fine-tuned using truncated texts that contain the minimum sufficient clues required to answer each question, and then applies this truncation to the reranked clues, thereby reducing reasoning tokens and improving the system’s \textbf{efficiency}.

Experiments on NQ, TriviaQA, and HotpotQA datasets show that CompSelect consistently outperforms baselines while requiring significantly less context for inference on both LLaMA3 and Qwen3. In addition to performance gains, it proves robust to unreliable retrieval and generalizes well across scenarios. By aligning retrieval with the reasoning needs of LLMs, CompSelect provides a scalable, cost-efficient solution for web-scale RAG applications.

The innovations in this paper are as follows:
\begin{itemize}
    \item We frame the LLM-centric compact clue selection mechanism as MinMax optimization, motivated by our finding that precise, compact, and well-organized inputs enhance both the LLM's reasoning performance and efficiency.
    \item We incorporate real feedback from the LLM: a KNN-based extractor gathers sufficient reasoning clues, a reranker organizes them into a coherent order, and a truncator improves reasoning density.
    \item Experiments on three datasets demonstrate that compact clue selection not only enhances inference performance and reduces system latency, but provides robustness against unreliable retrieval and supports generalization across tasks.
\end{itemize}

\begin{figure*}[!t]
\resizebox{0.8\textwidth}{!}{\includegraphics{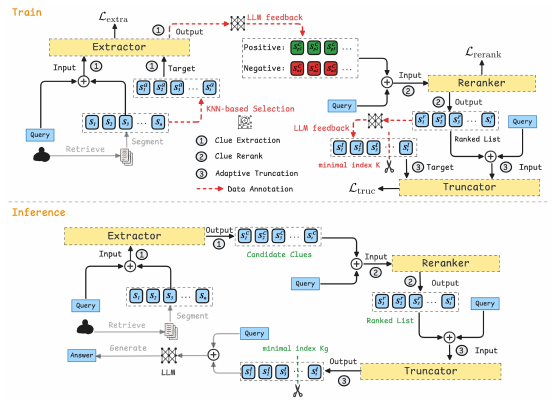}}
\centering
\caption{Architecture of CompSelect, consisting of three modules: a clue extractor, a reranker, and a truncator.}
\label{fig:architecture}
\end{figure*}

\section{Related Work}

Existing methods for identifying supporting content in RAG can be categorized as reranking or summarization, with summarization identifying query-relevant sentences via extractive selection or abstractive generation~\cite{selection}.

Reranking methods prioritize the most relevant content from retrieved documents. Classical methods like BM25~\cite{robertson2009probabilistic} rank documents based on term frequency, inverse document frequency, and document length. Neural rerankers~\cite{mitra2018introduction,mitra2017neural}, such as BGE-Reranker~\cite{bge} and RECOMP-extr~\cite{xu2024recomp}, use dense embeddings or fine-tuned cross-encoders to select salient sentences. RichRAG~\cite{wang-etal-2025-richrag} employs a generative list-wise ranker to produce and rank candidate documents, while ~\citet{ke-etal-2024-bridging} introduces a bridge mechanism to better connect retrievers and LLMs. CPC~\cite{liskavets2025prompt} ranks sentences with context-aware embeddings. Although reranking methods maintain low latency, they may harm RAG performance by disrupting the original logical structure of the document and can generate unfaithful clues.

Extractive methods aim to select key information from retrieved documents and can be further categorized into fragment selection and token pruning. Fragment selection methods, such as EXIT~\cite{exit}, use sentence-level classifiers to pick the most relevant sentences, achieving low latency but often relying on rigid selection criteria that may not adapt well to query complexity or document quality. Token-pruning methods, including LongLLMLingua~\cite{jiang2023longllmlingua} and Selective-Context~\cite{selective-context}, leverage LLMs to compute the informativeness of each token and remove low-information tokens, which may cause context fragmentation and higher latency. Other approaches, such as AdaComp~\cite{zhang2024adacomp}, BlendFilter~\cite{wang-etal-2024-blendfilter}, ClueAnchor~\cite{clueanchor} and EviOmni~\cite{zhao2025learning}, employ adaptive selection, knowledge blending, or reinforcement learning to improve reasoning efficiency and accuracy.

Abstractive methods generate query-relevant content via autoregressive generation. RECOMP-abs~\cite{xu2024recomp} uses a T5-based model~\cite{raffel2020exploring}, Refiner~\cite{refiner} hierarchically structures query-relevant content with larger LLMs, and BottleNeck~\cite{zhu2024information} uses reinforcement learning with information bottleneck theory to generate task-relevant summaries. FILCO~\cite{wang2023learning} trains a filtering model to dynamically select key sentences and jointly learns with the generator for end-to-end distillation. SEER~\cite{seer} uses self-alignment to select evidence via expert evaluation, but still depends on domain-specific experts and can be sensitive to noisy retrieval. These approaches reorganize scattered information into concise outputs, and as a result of their autoregressive generation, they typically incur higher latency.

In contrast, our work maximizes sufficiency, prioritizes effectiveness, and minimizes reasoning costs, enhancing both efficiency and accuracy in LLM-centric RAG.

\section{Problem Formulation}

Given a query $q$ and its retrieved document set $\mathcal{D} = \{d_1, \ldots, d_n\}$, where each document $d_i$ consists of a set of sentences $\mathcal{S}_i = \{s^i_1, \ldots, s^i_{n_i}\}$, with $n_i$ denoting the number of sentences in $d_i$.
The objective of our task is to identify an optimal subset $\mathcal{S}^* \subseteq \bigcup_{i=1}^{n} \mathcal{S}_i$ such that the language model $f_\theta$ generates the correct answer $y$ for the query $q$, while minimizing the subset size.
The optimal subset $\mathcal{S}^*$ can be determined by the following MinMax optimization:
\begin{align}
\mathcal{S}^* &= \arg\min\limits_{\mathcal{S'} \subseteq \mathcal{S}_{\text{max}}} |\mathcal{S'}|, \\
\mathcal{S}_{\text{max}} &= \arg\max\limits_{\mathcal{S} \subseteq \bigcup_{i=1}^n \mathcal{S}_i} \log P_\theta(y \mid \mathcal{S}, q),
\end{align}
where $\mathcal{S'}$ is the minimal subset of $\mathcal{S}_{\max}$ that still allows the model $f_\theta$ to generate the correct answer and $|\mathcal{S'}|$ denotes the number of sentences in $\mathcal{S'}$.
The selection of  \(\mathcal{S}^*\) should dynamically adapt to the real feedback of a RAG system to balance informativeness and conciseness. 

\section{Methodology}

In this section, we introduce CompSelect, a three-stage framework for selecting and prioritizing relevant information in RAG, as illustrated in Figure~\ref{fig:architecture}. CompSelect consists of three modules: a clue extractor, a clue reranker, and an adaptive truncator. The clue extractor identifies candidate answer clues from multiple documents, reducing the search space while retaining relevant information. The reranker then prioritizes the most relevant clues using a pairwise loss strategy. Finally, the truncator adaptively selects the minimal necessary context for each query, thereby improving inference efficiency and answer accuracy~\footnote{Our code is available at~\url{https://github.com/zqc1023/CompSelect}.}.

\subsection{Clue Extractor}

The goal of the clue extractor is to identify potential answer clues from multiple documents and construct a smaller query-relevant candidate set to reduce the search space. We compare the performance of answer-containing sentences and the original retrieved documents in downstream tasks, as shown in Figure~\ref{fig:figure1}. The results indicate that filtering out low-value information from documents benefits RAG reasoning. Although not all answer-containing sentences are query-relevant, they approximate the maximal subset capable of addressing user queries, and can serve as the optimization target for the clue extractor.

To guide the extraction of informative sentences, we first identify sentences from the retrieved documents that contain the ground-truth answer as targets for extraction.
Given a query $q$ and the ground-truth answer $y$, we denote by $\mathcal{S} = \{s_1, \dots, s_n\}$ the set of sentences appearing in the retrieved documents, with sentences preserving their original order within and across documents. The answer-containing sentences are defined as:
\begin{align}
    \mathcal{S}^{a}= \{s_j|y \sqsubseteq s_j, s_j \in \mathcal{S} \},
\end{align}
where $y \sqsubseteq s_j$ indicates that $y$ is a substring of $s_j$.

Next, we fine-tune an LLM as the clue $Extractor$ to generate answer-containing sentences $ \mathcal{S}^{a}$ based on the query $q$ and the retrieved sentences $\mathcal{S}$ with a specific prompt (see Appendix~\ref{app:prompt-extr}). The loss function of $Extractor$ model is defined as:
\begin{align}
\mathcal{L}_\text{extra} = - \frac{1}{N} \sum_{i=1}^{N} \log P_{\theta}(\mathcal{S}^{a}_i \mid q_i, \mathcal{S}_i).
\end{align}

Finally, the clue extractor is capable of generating the potential candidate clues based on the user query and retrieved sentences during inference:
\begin{align}
    \mathcal{S}^{c} = Extractor(q,\mathcal{S}).
\end{align}

This procedure enables the RAG system to focus on informative content, thereby improving efficiency and answer accuracy.

\begin{table}[!h]
\centering
\renewcommand{\arraystretch}{1.1}
\caption{SubEM (\%) comparison of different strategies on NQ, TriviaQA, and HotpotQA test set. We use LLaMA3-8B-Instruct as the generator. $\epsilon$ shows the KNN threshold and higher values introduce more contextual sentences.}
\resizebox{0.9\columnwidth}{!}{
\begin{tabular}{lccc}
\toprule[1.1pt]
\textbf{Method }& \textbf{NQ} & \textbf{TriviaQA} & \textbf{HotpotQA} \\
\midrule
Full Content                    & 45.87               & 67.08 & 25.66 \\
Sentences \textit{w/} answer        & \textbf{56.73} & 71.52 & 30.46 \\
KNN-based Extraction ($\epsilon = 0.15$)           & 56.41                  & \textbf{71.98} & \textbf{30.63}\\
\bottomrule[1.1pt]
\end{tabular}%
}
\label{tab:subem_comparison_transposed}
\end{table}

\subsubsection{\textbf{KNN-based Extraction}} 
We further augment the clue extractor with a KNN-based strategy.
As shown in Table~\ref{tab:subem_comparison_transposed}, selecting answer-containing sentences substantially improves performance on single-hop QA datasets such as NQ, raising the SubEM score from 45.87 to 56.73. This highlights the benefit of focusing on relevant context for simple queries.
For multi-hop or more complex QA datasets (e.g., TriviaQA and HotpotQA), incorporating answer-containing sentences also leads to performance gains. Moreover, further retrieving semantically similar sentences using a KNN-based strategy yields additional improvements, indicating that extra contextual information beyond strictly answer-containing sentences is beneficial for multi-hop reasoning and complex query understanding.
Based on these observations, we propose a KNN-based extraction strategy. For simple questions, we select answer-containing sentences as optimization targets to provide key information. For complex questions, we additionally retrieve semantically similar sentences to supplement context. 
By using both answer-containing and KNN-based similar sentences as optimization targets, the extractor adapts to queries of varying complexity, improving reasoning accuracy while maintaining a concise candidate set.

\subsection{Clue Reranker}
The clue extractor often selects sentences with multiple relevant clues of varying importance, necessitating reranking. We address this by training a reranker with pairwise loss to prioritize the most relevant sentences.

\subsubsection{\textbf{Training}} We use real downstream generator feedback from the RAG system to annotate the training data for the reranker, as QA performance on complex questions heavily depends on the characteristics of the generation module. 
First, we pair each of the extracted clue sentences $s_j^c \in \mathcal{S}^{c}$ with the query $q$ as $(s_j^c,q)$, where sentence $s_j^c$ that enables the downstream generation module to produce the correct answer for $q$ is considered as positive sample $s_{\text{pos}}$, while other sentences are treated as negative samples $s_{\text{neg}}$\footnote{If no candidate clues can generate the correct answer, or if all samples can generate the correct answer, the sample will be removed from the annotated data.}. The $Reranker$ aims to minimize the following pairwise loss function~\cite{karpukhin2020dense} to improve relevance ranking:

\begin{align}
\mathcal{L}_{\text{rerank}} = - \frac{1}{N} \sum_{i=1}^{N} \log \frac{e^{\text{sim}(q_i, s_{\text{pos}}^{i})}}{e^{\text{sim}(q_i, s_{\text{pos}}^{i})} + e^{\text{sim}(q_i, s_{\text{neg}}^{i})}},
\end{align}

where \(\text{sim}(q, *)\) represents the semantic similarity between the query \(q\) and the sentence $*$ by $Reranker$ model. Minimizing this loss function enables the $Reranker$ model to effectively identify and prioritize the most relevant clues. 
\subsubsection{\textbf{Inference}}
Given the query $q$ and the extracted sentences $\mathcal{S}^{c}$, the $Reranker$ model calculates the relevance score between every sentence $s_j^c\in \mathcal{S}^{c}$ and query $q$. The reranked clue set is defined as:
\begin{align}
     \mathcal{S}^{r} = Reranker(q,\mathcal{S}^{c}).
\end{align}

\subsection{Adaptive Truncator}
The adaptive truncator is designed to identify the minimal necessary clues based on the complexity of the question and the documents, ensuring sufficient clues for accurate answer generation.

\subsubsection{\textbf{Training}}  To determine the optimal clue subset $\mathcal{S}^{t}$ for each query $q$, we perform data annotation based on the reranked answer clues $\mathcal{S}^{r}$ obtained from the previous reranking step. Given a query $q$ and its reranked clues $\mathcal{S}^{r} =  \{s_1^r, \dots, s_n^r\}$, the objective is to identify the smallest subset $\mathcal{S}^{t}$ such that the RAG system's generation model $M$ can generate the correct answer $y$ based on $q$ and $\mathcal{S}^{t}$. We define $D_k = \{s_1^r, \dots, s_k^r\}$, where $1 \leq k \leq n$. The performance on each subset  $D_k$ is evaluated by checking if the generation model's output $M(q,D_k)$ matches the ground truth \( y \). The correctness condition is defined as:
\begin{equation}
\text{Correct}(q, D_k) = 
\begin{cases}
1, & \text{if } M(q, D_k) = y \\
0, & \text{otherwise}
\end{cases}.
\end{equation}

Since the reranker cannot guarantee that the most relevant sentences are always ranked first, especially for complex questions, we iterate over the subsets from largest to smallest, starting with \( D_n \) and continuing to \( D_1 \). The optimal subset $\mathcal{S}^{t}$ is the smallest subset that generates the correct answer:
\begin{align}
    \mathcal{S}^{t} &= \{s^r_1,\dots,s^r_K \},\\
     K &= \arg \min_k \{ k \mid \text{Correct}(q, D_k) = 1 \}.
\end{align}

If the RAG system cannot generate a correct answer from any subset, then \(\mathcal{S}^{t} = \emptyset \), indicating no subset suffices. This method ensures the use of minimal necessary context \(\mathcal{S}^{t} \).

During the model training stage, we fine-tune an LLM  based on the data annotations. The $Truncator$ is trained to predict the minimal subset of reranked clues $\mathcal{S}^{t} \subseteq \mathcal{S}^{r}$ that suffices to answer the query:
\begin{align}
   \mathcal{L}_\text{truc} &= -\frac{1}{N} \sum_{i=1}^{N} \log P_{\theta}(\mathcal{S}^{t}_i | q_i, \mathcal{S}^{r}_i).
\end{align}

\subsubsection{\textbf{Inference}}
During inference, given a new query $q$ and its reranked sentences $\mathcal{S}^{r}$, the $Truncator$ truncates $\mathcal{S}^{r}$ to the minimal sufficient clue set $\mathcal{S}^{t} = \{s^r_1, \ldots, s^r_{K_g}\}$, where $K_g$ represents the number of sentences needed to answer the query. The truncation process uses the prompt provided in Appendix~\ref{app:prompt-trunc}. 
Finally, the RAG generation module concatenates the query $q$ with the filtered answer clues $\mathcal{S}^{t}$ using the following prompt to reason the answer:  

\begin{table}[H]
\begin{tcolorbox}
\verb|<system>|\\
You are a helpful, respectful, and honest assistant. Answer the question with a couple of words using the provided documents. For example: Question: What is the capital of France? Output: Paris.\\
\verb|</system>|
\\
\verb|<user>|\\
Question: \verb|{query}|\\
Documents: \\
Doc1: \verb|{Document 1}|  \\
Doc2: \verb|{Document 2}|  \\
...... \\
\verb|</user>|\\
\end{tcolorbox}

\label{tab:prompt3}
\end{table}

\section{Experiments}

\begin{table*}[!t]
  \centering
   \setlength\tabcolsep{14pt}
\renewcommand\arraystretch{1.1}
    \caption{ Experimental results (\%) on three datasets using LLaMA3-8B-Instruct and Qwen3-14B as the generators. The retriever is DPR. All baselines and our method are conducted using the same test sets and retrieval corpus.}
\resizebox{0.95\textwidth}{!}{
\scalebox{0.4}{
  \begin{tabular}{l ccccccccc}
     \toprule[1.1pt]
\multirow{2}{*}{\textbf{Method}} & \multicolumn{3}{c}{\textbf{NQ}} &\multicolumn{3}{c}{\textbf{TriviaQA}} & \multicolumn{3}{c}{\textbf{HotpotQA}}   \\
 \cmidrule(lr){2-4} \cmidrule(lr){5-7} \cmidrule(lr){8-10}
    &  \textbf{SubEM}  & \textbf{F1}  & \textbf{ CR}  & \textbf{SubEM} & \textbf{F1} &\textbf{CR}  & \textbf{SubEM}  & \textbf{F1} & \textbf{CR} \\
    \hline
          \rowcolor[HTML]{F0F0F0}
   \multicolumn{10}{c}{\textbf{\mbox{\textsc{LLaMA3-8B-Instruct}}\xspace}} \\

Naive Generation & 25.18 & 29.11 & -  & 55.92 & 58.95  & - & 21.39 & 22.87  & - \\
Full Content & 45.87 & 47.86 & 1$\times$  & 67.08 & 68.61 & 1$\times$ & 25.66 & 28.22 & 1$\times$  \\

\cdashline{1-10}

  BM25  & 38.97 & 41.21 & 4.1$\times$  & 53.17  & 56.73 & 4.3$\times$ & 21.85 & 23.14 & 4.4$\times$\\
  Bge-reranker  & 41.22 & 45.56 & 4.1$\times$ & 57.93  & 60.43 & 4.3$\times$ & 22.73 & 24.01 & 4.5$\times$  \\  
  
  RECOMP-extr & 40.71 & 45.35 & 11.97$\times$ & 63.73 & 66.46  &10.91$\times$ & 24.39 & 26.54 & 8.33 $\times$ \\

    \cdashline{1-10}
  LongLLMLingua  & 41.85 & 45.74 & 4.56$\times$ & 64.92 & 66.87 & 4.18$\times$ & 23.74 & 26.19 & 4.45$\times$  \\
 
    Selective-Context & 43.84 & 46.33 & 2.6$\times$ & 61.53 & 61.02 & 2.7$\times$ & 24.28 & 26.51 & 2.7$\times$  \\
     EXIT  & 41.19 & 45.44 & 14.16$\times$ & 64.03 & 66.46 & 12.78$\times$ & 24.16 & 25.80 & \textbf{15.43$\times$}  \\
    
  \cdashline{1-10}
  RECOMP-abs & 43.11 & 46.16 & 11.12$\times$ & 61.89 & 61.35 & 11.25$\times$ & 23.55 & 25.29 & 7.91$\times$ \\
  Refiner & 46.12 & 48.37 & 10.97$\times$ & 65.97 & 67.64 & 12.63$\times$ & 24.72 & 26.78 & 7.65$\times$  \\
  BottleNeck & 45.32 & 47.21 & 14.32$\times$ & 66.98 & 68.31 & \textbf{13.17$\times$} & 25.71 & 28.13 &13.21$\times$ \\
      
\cdashline{1-10}
  \textbf{Ours ($\epsilon = 0$)} & \textbf{46.98} & \textbf{49.45} & \textbf{14.95$\times$} & \textbf{68.29} & \textbf{69.72} & 12.54$\times$ & \textbf{26.14} & \textbf{28.43} & 13.56$\times$  \\
    \hline
      \rowcolor[HTML]{F0F0F0}
     \multicolumn{10}{c}{\textbf{\textsc{Qwen3-14B}}} \\
          
Naive Generation & 27.92 & 29.01 & -  & 56.56 & 57.71  & - & 23.59 & 29.51 & - \\
Full Content & 51.52 & 48.55 & 1$\times$  & 72.67 & 72.10 & 1$\times$  & 30.05 & 34.32 & 1$\times$  \\
\cdashline{1-10}

  BM25  & 40.88 & 45.52 & 4.1$\times$ & 61.15  & 60.91 & 4.3$\times$ & 24.08 & 26.31 & 4.4$\times$ \\
  Bge-reranker  & 42.95 & 46.15 & 4.1$\times$ & 63.12  & 66.57 & 4.3$\times$ & 24.73 & 26.79 & 4.5$\times$  \\  
  RECOMP-extr & 43.29 & 46.25 & 11.97$\times$  & 62.94 & 63.86 & 10.91$\times$  & 25.77 & 29.83 & 8.33$\times$  \\

  \cdashline{1-10}
  LongLLMLingua  & 44.79 & 46.93 & 4.56$\times$ & 68.73 & 68.39 & 4.18$\times$ & 26.43 & 30.21 & 4.45$\times$  \\

    Selective-Context & 49.31 & 47.22 & 2.6$\times$ & 65.59 & 67.11 & 2.7$\times$ & 26.07 & 30.05 & 2.7$\times$  \\
      EXIT  & 42.23 & 45.77 & 14.16$\times$ & 63.78 & 64.85 & 12.78$\times$ & 28.16& 33.89 & \textbf{15.43$\times$} \\

  \cdashline{1-10}
    
 RECOMP-abs & 45.75 & 47.56 & 11.12$\times$ & 65.34 & 66.93 & 11.25$\times$ & 27.45 & 31.19 & 7.91$\times$ \\
   Refiner  & 51.14 & 48.16 & 10.97$\times$ & 71.18 & 70.55 & 12.63$\times$ & 30.12 & 34.39 & 7.65$\times$  \\
  BottleNeck & 50.72 & 47.78 & 14.32$\times$ & 72.36 & 71.87 & 13.17$\times$ & 29.64 & 33.72 & 13.21$\times$ \\
      
\cdashline{1-10}
 \textbf{Ours ($\epsilon = 0$)} & \textbf{52.33} & \textbf{49.32} & \textbf{16.18$\times$} & \textbf{72.93} & \textbf{72.61} & \textbf{13.35$\times$} & \textbf{30.67} & \textbf{34.61} & 14.17$\times$  \\
      \bottomrule[1.1pt]
      \end{tabular} 
  }
    }
 
  \label{tab:main_results1}
\end{table*}

\subsection{Experimental Setup}
\subsubsection{\textbf{Datasets}} 

We evaluate our method on three QA datasets, including (1) Open-Domain QA, represented by NaturalQuestions (NQ)~\cite{kwiatkowski2019natural} and TriviaQA~\cite{joshi2017triviaqa}; (2) Multi-Hop QA, represented by HotpotQA~\cite{yang2018hotpotqa}. 
Table~\ref{qa-datasets} in Appendix~\ref{Appendix:dataset} illustrates the statistics of these datasets.

\subsubsection{\textbf{Evaluation Metrics}}
Since answer style mismatch may cause additional variance, we follow prior work~\cite{zhu-etal-2024-atm,xu2024search,wang2025maferw} and adopt Substring Exact Match (\textbf{SubEM}) and \textbf{F1} for evaluation. SubEM checks whether the gold answer appears as a substring in the prediction, while F1 measures token-level overlap with the reference. 
For efficiency, we report the compression ratio (\textbf{CR}), defined as the ratio of original to compressed context length. We also measure \textbf{Total Latency}, including offline preprocessing and online answer generation, and \textbf{Online Latency}, the time from user query submission to answer generation using the preprocessed results, which can provide a more comprehensive assessment of practical efficiency.

\subsubsection{\textbf{Baselines}} We consider four baseline strategies: 

\textbf{Vanilla Methods:}
(i) \textbf{\textit{Naive Generation}}, which relies solely on the generator’s parametric knowledge;
(ii) \textbf{\textit{Full Content}}, which concatenates all the retrieved documents as input.

\textbf{Reranking Methods:}
(i) \textbf{\textit{BM25}}~\cite{robertson2009probabilistic}, a classic lexical matching method that scores and ranks documents based on term frequency, inverse document frequency, and document length normalization;  
(ii) \textbf{\textit{BGE-reranker}}~\cite{bge}, a neural reranker that computes dense embeddings for queries and documents, ranking documents according to semantic similarity in the embedding space;  
(iii) \textbf{\textit{RECOMP-extr}}~\cite{xu2024recomp}, which employs a fine-tuned cross-encoder to select salient sentences through dense retrieval.

\textbf{Extractive Methods:}
(i) \textbf{\textit{LongLLMLingua}}~\cite{jiang2023longllmlingua}, which prunes irrelevant tokens in long contexts via a dynamic programming algorithm guided by question-aware perplexity scores;
(ii) \textbf{Selective-Context}~\cite{selective-context}, which uses self-information estimated by an external LLM to prune redundant words;
(iii) \textbf{\textit{EXIT}}~\cite{exit}, which compresses retrieved documents by applying sentence-level relevance classification to select and reassemble only the most relevant sentences for answering queries.

\textbf{Abstractive Methods:}
(i) \textbf{\textit{RECOMP-abs}}~\cite{xu2024recomp}, which uses a T5-based model to perform abstractive summarization, compressing documents into shorter token sequences through autoregressive generation;
(ii) \textbf{\textit{Refiner}}~\cite{refiner},  which leverages large LLMs to extract and structure query-relevant content from retrieved documents, producing a hierarchical output based on intrinsic document knowledge;
(iii) \textbf{\textit{BottleNeck}}~\cite{zhu2024information}, which employs reinforcement learning and information bottleneck theory to improve both filtering and generation.

\subsubsection{\textbf{Implementation Details}}
\label{imple}
We use LLaMA3-8B-Instruct~~\cite{llama3}, Qwen3-14B~\cite{qwen3}, and LLaMA3.3-70B-Instruct~\cite{llama3} (presented in Appendix~\ref{Appendix:70b}) as the generators,  covering medium, large, and ultra-large LLMs. To ensure high coverage and quality of retrieved information~\cite{cuconasu2024power}, we follow prior work~\cite{zhu2024information,xu2024recomp,refiner,zhang2026stable} and utilize the adversarial Dense Passage Retriever (DPR)~\cite{karpukhin2020dense} to retrieve Top-5 passages from the full Wikipedia passages for each query. All baselines and our method are conducted using the same test sets and retrieval corpus, which guarantees consistency between baseline methods and our approach. To further reduce computational cost and latency, our clue extractor and clue truncator are based on LLaMA3.2-3B-Instruct~\cite{llama3} and we train the model using the LoRA method~\cite{hu2021lora} within the LLaMAFactory framework~\cite{llamafactory}. For the clue reranker, we implement Sentence-BERT~\cite{reimers2020making} using distilbert-base-uncased. More details are provided in Appendix~\ref{Appendix:settings}.

\subsection{Main Results}
The comparison results for the three datasets are shown in Table~\ref{tab:main_results1}. The results indicate the following: 
(i) \textbf{RAG improves downstream task performance across all datasets.} \quad Incorporating retrieved documents consistently boosts answer accuracy compared with using the generator alone;
(ii) \textbf{Compact clue selection improves information utilization.}\quad Compared with Full Content, our method significantly compresses the context (high CR) while maintaining or improving SubEM and F1, effectively reducing redundant information and retaining critical clues for accurate reasoning;
(iii) \textbf{CompSelect consistently achieves the best overall performance.}\quad Across different datasets and generators, our method attains the highest SubEM and F1 scores, highlighting its effectiveness in extracting key information and generating accurate answers; 
(iv) \textbf{CompSelect maintains robust performance and generalizes well across datasets and models.} \quad  Across multiple datasets and generators, our method consistently outperforms the baselines, indicating stable performance and strong generalization in diverse QA scenarios.

\begin{figure*}[t]
\centering
\includegraphics[width=0.95\textwidth]{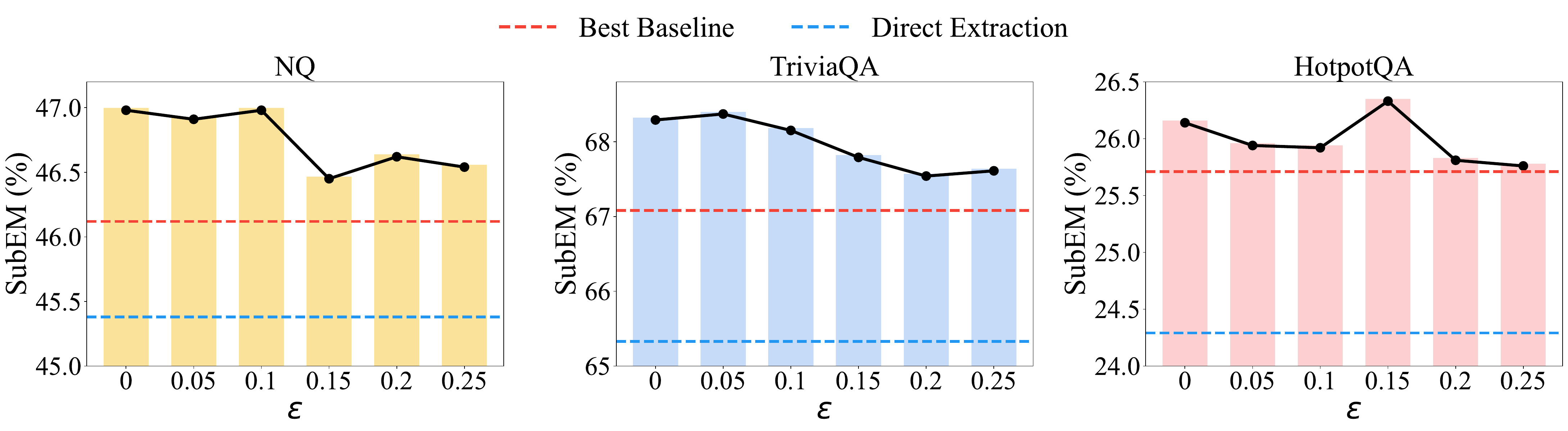} 
\caption{An illustration of clue extraction performance using LLaMA3-8B-Instruct as generator on NQ, TriviaQA, and HotpotQA test sets. The x-axis shows the KNN threshold and higher values introduce more contextual sentences.}
\label{fig:performance_compare}
\end{figure*}

\subsection{Ablation Study}
\subsubsection{\textbf{Overall.}}
To explore the impact of different components of CompSelect, we use LLaMA3-8B-Instruct as the base LLM and introduce the following variants for ablation study:  
1) \textit{w/o clue extractor.} This variant directly uses LLaMA3.2-3B-Instruct without fine-tuning for clue extraction.  
2) \textit{w/o clue reranker.} This variant skips reranking and retains the original sentence order.  
3) \textit{w/o adaptive truncator.} This variant disables adaptive truncation of the reranked clues.  
As shown in Table~\ref{tab:2}, removing any single component leads to a drop in SubEM, indicating that each module contributes significantly to the overall performance.

\begin{table}[t]
  \centering
  \renewcommand\arraystretch{1.1}
  \small
   \caption{Ablation study on NQ, TriviaQA and HotpotQA test set. 
We use LLaMA3-8B-Instruct as the generator and SubEM (\%) for evaluation.}

   \resizebox{0.45\textwidth}{!}{ 
  
  \begin{tabular}{l c c c}
   \toprule[1.1pt]
\multirow{1}{*}{\textbf{Method}}    &   \multicolumn{1}{c}{\textbf{NQ}} &\multicolumn{1}{c}{\textbf{TriviaQA}} & \multicolumn{1}{c}{\textbf{HotpotQA}} \\
    \midrule
    \textbf{CompSelect (Ours)}  & \textbf{46.98}  & \textbf{68.29}  & \textbf{26.14}\\
    \cdashline{1-4}
     \multicolumn{1}{l}{\quad \textit{w/o clue extractor}} & 45.83 & 66.71 & 25.12 \\
     \multicolumn{1}{l}{\quad \textit{w/o clue reranker}} & 46.55& 67.95 & 25.84 \\
     \multicolumn{1}{l}{\quad \textit{w/o adaptive truncator}} & 46.72 & 67.89 & 25.91 \\
     \bottomrule[1.1pt]
  \end{tabular}
  }
 
  \label{tab:2}
\end{table}

\subsubsection{\textbf{Direct vs. Fine-tuned Extraction}}
We analyze the effect of fine-tuning on clue extraction by comparing two approaches: 1) \textit{Direct Extraction}, which uses the extractor without fine-tuning to extract clue sentences from retrieved documents based on the given prompt (see Appendix~\ref{app:prompt-extr}); 2) \textit{Fine-tuned Extraction}, which uses fine-tuned extractor to select answer-containing sentences.
As shown in Table~\ref{tab:finetune}, Fine-tuned Extraction consistently outperforms Direct Extraction by leveraging task-specific knowledge to identify more relevant sentences.
\begin{table}[h] 
    \centering
    \renewcommand\arraystretch{1.0}
     \small
      \caption{Performance (\%) comparison between Direct Extraction and Fine-tuned Extraction across three datasets. We use LLaMA3-8B-Instruct as the generator. }
         \resizebox{0.45\textwidth}{!}{ 
    \begin{tabular}{lccc}
        \toprule[1.1pt]
         \multicolumn{1}{l}{\textbf{Dataset}}  &\textbf{Method} & \textbf{SubEM} & \textbf{F1} \\ 
        \midrule
         \multirow{2}{*}{NQ}   &Direct Extraction & 45.38& 47.27 \\
         &\textbf{Fine-tuned Extraction} & \textbf{46.43} & \textbf{49.45}\\
        \cdashline{1-4}
       \multirow{2}{*}{TriviaQA} &Direct Extraction& 65.33  & 68.28 \\
       &  \textbf{Fine-tuned Extraction} &\textbf{67.54}  & \textbf{69.72}\\
        \cdashline{1-4}
        \multirow{2}{*}{HotpotQA}&Direct Extraction & 24.29 & 25.74\\
       &  \textbf{Fine-tuned Extraction} & \textbf{25.84} & \textbf{28.50}\\
       \bottomrule[1.1pt]
    \end{tabular}
    }
   
    \label{tab:finetune}

\end{table}

\subsubsection{\textbf{Threshold of KNN-Based Extraction}}

We assess the KNN-based extraction method by varying the threshold, which controls the cosine similarity to answer-containing sentences. A threshold of 0 selects only the answer sentences, while higher thresholds allow semantically similar sentences to expand the context with additional relevant information.
As shown in Figure~\ref{fig:performance_compare}, we compare the model's performance at different threshold values. For simple questions such as NQ, the KNN extraction strategy does not improve performance, as answers can typically be obtained directly from sentences containing the answers. For more complex questions such as HotpotQA and TriviaQA, the KNN strategy improves performance at lower thresholds but declines at higher thresholds due to increased noise. Notably, KNN-based extraction serves a supplementary role in the overall framework.
While the threshold setting impacts CompSelect’s performance, it does not alter the experimental finding that CompSelect outperforms the best baseline.

\subsubsection{\textbf{Random vs. Adaptive Truncation}}
To validate the effectiveness of our adaptive truncation strategy, we compare it with random truncation. As shown in Table~\ref{tab:random}, our adaptive truncation consistently outperforms random truncation in both SubEM and F1 scores. This improvement arises because adaptive truncation dynamically selects the most relevant context based on real-time feedback from the downstream generator, retaining sentences that are both highly informative and directly pertinent to the query. By doing so, it preserves critical information while reducing unnecessary content, enhancing the model’s answer accuracy and overall reasoning capability. In contrast, random truncation does not account for the importance of individual sentences and may discard key clues, leading to degraded performance.
\begin{table}[H] 
    \centering
    \renewcommand\arraystretch{1.0}
     \small
      \caption{Performance (\%) comparison between Random Truncation and Adaptive Truncation across three datasets. We use Llama3-8B-Instruct as the generator. We report SubEM and F1 for evaluation. }
         \resizebox{0.45\textwidth}{!}{ 
    \begin{tabular}{lccc}
        \toprule[1.1pt]
         \multicolumn{1}{l}{\textbf{Dataset}}  &\textbf{Method} & \textbf{SubEM} & \textbf{F1} \\ 
        \midrule
         \multirow{2}{*}{NQ}   &Random & 46.35& 48.44 \\
         &\textbf{Adaptive Truncation} & \textbf{46.98} & \textbf{49.45}\\
        \cdashline{1-4}
       \multirow{2}{*}{TriviaQA} &Random& 67.71  & 69.13 \\
       &  \textbf{Adaptive Truncation} &\textbf{68.29}  & \textbf{69.72}\\
        \cdashline{1-4}
        \multirow{2}{*}{HotpotQA}&Random & 25.73 & 28.17\\
       &  \textbf{Adaptive Truncation} & \textbf{26.14} & \textbf{28.43}\\
       \bottomrule[1.1pt]
    \end{tabular}
     }
    \label{tab:random}
\end{table}

\begin{figure}[!t]
\centering
  \includegraphics[width=0.95\columnwidth]{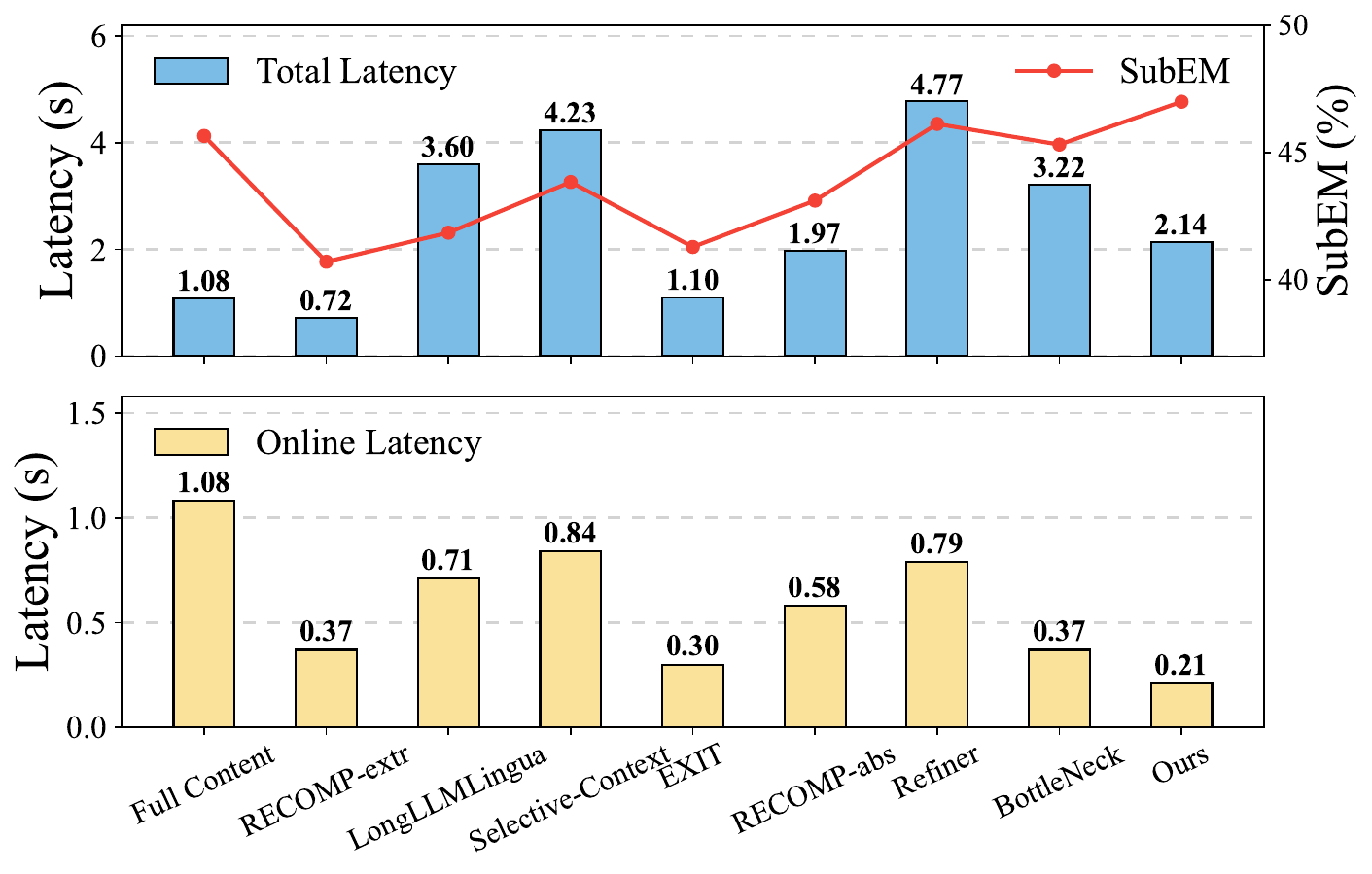}
\caption{Latency comparison across baselines and our method using LLaMA3-8B-Instruct. The top shows Total Latency along with SubEM performance, while the bottom shows Online Latency. Experiments were conducted on two NVIDIA RTX PRO 6000 GPUs.}
  \label{time}
\end{figure}
\subsection{\textbf{System Latency Evaluation}}

To evaluate the overall efficiency of each method, we measured the average total latency (including offline preprocessing and online answer generation) and online latency (the time to answer generation using the preprocessed results) for processing a single sample on the NQ test set, along with the corresponding SubEM scores. This measurement setup provides a more comprehensive assessment of practical efficiency. As shown in Figure~\ref{time}, RECOMP-extr exhibits the lowest total latency but relatively lower SubEM, whereas our method achieves the highest SubEM while maintaining a relatively low total latency and online latency. This indicates that with lightweight reranker and truncator design, our method can improve accuracy without significantly increasing latency, thereby achieving a better trade-off between efficiency and accuracy.

\subsection{\textbf{Robustness Analysis}}
\label{app:robustness}
\subsubsection{\textbf{Cascading Errors Resilience Analysis}}

To quantify cascading errors, we conduct experiments on the three test sets in which the retrieved documents contain the gold-standard answers. Specifically, \textbf{Recall-1} evaluates whether the extractor successfully recalls the gold-standard answer, \textbf{Hit-2@1} measures the probability that the reranker ranks the gold-standard answer as Top-1, and \textbf{Recall-3} assesses whether the truncator continues to retain the gold-standard answer after truncation. These metrics provide a systematic means to analyze how errors propagate across the sequential components of the system. As shown in Table~\ref{tab:cascade_errors}, CompSelect consistently maintains a low error rate across all modules, indicating that cascading errors are effectively controlled and kept within a manageable range, which demonstrates the robustness of the framework in preserving critical information throughout the  processing pipeline.

\begin{table}[t]
  \centering
  \small
  \renewcommand{\arraystretch}{1.0}
   \caption{Cascading errors analysis of extractor (Recall-1), reranker (Hit-2@1) and truncator (Recall-3) on three datasets, with $\epsilon$ set to 0.}

  \begin{tabular}{lccc}
    \toprule[1.1pt]
    \textbf{Dataset} & \textbf{Recall-1 (\%)} & \textbf{Hit-2@1 (\%)} & \textbf{Recall-3 (\%)} \\
    \midrule
    NQ         & 80.71 & 78.35 & 79.46 \\
    TriviaQA   & 92.33 & 84.46 & 86.07 \\
    HotpotQA   & 84.12 & 77.59 & 80.16 \\
    \bottomrule[1.1pt]
  \end{tabular}
 
  \label{tab:cascade_errors}
\end{table}

\begin{table}[t]
  \centering
  \small
  \renewcommand{\arraystretch}{1.0}
   \caption{Results under unreliable retrieval. We use LLaMA3-8B-Instruct as the generator and F1 (\%) for evaluation. The best results are in bold and the second are
\underline{underlined}.}
      \begin{tabular}{llcc}
    \toprule[1.1pt]
    \multicolumn{2}{l}{\textbf{Method}}&\textbf{NQ} &\textbf{HotpotQA} \\
    \midrule
     \multicolumn{2}{l}{RECOMP-extr}      &     14.93     &15.65\\
     \multicolumn{2}{l}{LongLLMLingua}     &   14.77   &15.77\\
     \multicolumn{2}{l}{Selective-Context}  &     \textbf{15.42}          &\underline{16.35}\\
    \multicolumn{2}{l}{EXIT}                &  14.59 &15.25\\
     \multicolumn{2}{l}{RECOMP-abs}           &   13.74   &15.08\\
       \multicolumn{2}{l}{Refiner}               &  15.13 &16.26\\
       \multicolumn{2}{l}{BottleNeck}               &  14.58 &16.18\\

    \multicolumn{2}{l}{\textbf{Ours (\textit{w/o Truncator})}} & 14.32 &16.12\\
    \rowcolor[HTML]{F0F0F0}
    \multicolumn{2}{l}{\textbf{Ours (\textit{w/ Truncator})}}   &  \underline{15.36}    &\textbf{16.87}\\ 
    \bottomrule[1.1pt]
  \end{tabular}
  \label{tab:robustness}
\end{table}
\subsubsection{\textbf{Performance under Unreliable Retrieval}} The truncator not only shortens context but also helps filter out unreliable content. We conduct experiments on samples from the NQ and HotpotQA test sets without gold-standard answers to simulate unreliable retrieval. As shown in Table~\ref{tab:robustness}, CompSelect's truncator effectively suppresses noise and prevents answer degradation. This is because CompSelect is trained to allow empty outputs, whereas baseline models perform poorly under unreliable retrieval conditions, as they must choose an output answer clue.

\subsection{\textbf{ Cross-Task Generalization}}

To further evaluate the generalization capability of CompSelect, we perform inference on 1,200 randomly sampled instances from a Conversational Multi-Doc QA dataset~\footnote{\url{https://sites.google.com/view/wsdm24-docqa}}, with the model trained solely on HotpotQA dataset. This setup allows us to examine whether the model can adapt to a new conversational QA scenario without direct exposure during training. As shown in Table~\ref{tab:generalization}, CompSelect consistently achieves strong performance in this unseen dataset, highlighting its robustness and ability to generalize across diverse QA tasks.

\begin{table}[t]
\centering
\renewcommand{\arraystretch}{1.1}
\small
\caption{Comparison of cross-task generalization ability across different baselines. We use ROUGE-1, ROUGE-2, and ROUGE-L for evaluation.}
\begin{tabular}{lccc}
\toprule[1.1pt]
\textbf{Method} & \textbf{ROUGE-1} & \textbf{ROUGE-2} & \textbf{ROUGE-L} \\
\midrule
Full Content     & 0.3769          & 0.1546          & 0.2234          \\
RECOMP-extr      & 0.3051          & 0.1022          & 0.1668          \\
LongLLMLingua    & 0.4082          & 0.1775          & 0.2369          \\
Selective-Context& 0.3927          & 0.1611          & 0.2275          \\
EXIT             & 0.3842          & 0.1573          & 0.2219          \\
RECOMP-abs        & 0.3795          & 0.1498          & 0.2146          \\
Refiner          & 0.4018          & 0.1729          & 0.2336          \\
BottleNeck       & 0.3708          & 0.1529          & 0.2158          \\
\rowcolor[HTML]{F0F0F0}
\textbf{Ours}    & \textbf{0.4123} & \textbf{0.1798} & \textbf{0.2512} \\
\bottomrule[1.1pt]
\end{tabular}
\label{tab:generalization}
\end{table}

\section{Conclusion}
In this work, we propose CompSelect, a compact clue selection mechanism for LLM-centric RAG that efficiently extracts, organizes, and truncates answer-relevant information from large-scale documents. By framing retrieval as a MinMax optimization, its three modules, the clue extractor, the reranker, and the truncator, work together to enhance reasoning quality and efficiency. Experiments on three QA datasets show that CompSelect improves performance and reduces latency, while remaining robust to unreliable retrieval and generalizing well across scenarios. Future work could explore integrating it with generative retrieval, allowing models to generate document indices, thereby avoiding costly online processing and further reducing end-to-end latency.

\section*{Acknowledgements}
This work was funded by the National Natural Science Foundation of China (NSFC) under Grants No.U25B2070 and No. 62406013, the Beijing Advanced Innovation Center Funds for Future Blockchain and Privacy Computing(GJJ-24-034), and the Fundamental Research Funds for the Central Universities.

\clearpage

\bibliographystyle{ACM-Reference-Format}
\bibliography{sample-base}

\appendix

\section{Prompt}
\subsection{Prompt for the Clue Extractor}
\label{app:prompt-extr}
We present our prompt for the clue extractor in Table~\ref{tab:prompt1}.
The prompt is designed to guide the model in extracting the most informative sentences, which are most likely to contain the answer to the given question.

\subsection{Prompt for the Adaptive Truncator}
\label{app:prompt-trunc}
We present our prompt for the adaptive truncator in Table~\ref{tab:prompt2}. The prompt is designed to guide the model in optimizing context truncation based on the complexity of the question and the quality of the retrieved documents, thereby improving the efficiency of the language model. 
Specifically, given a question and a ranked list of sentences, the model’s task is to identify and retain the most relevant sentences while truncating those that are irrelevant to the question. 

\begin{table}[H]
\caption{Prompt for the Clue Extractor.}
\begin{tcolorbox}
You are a highly skilled assistant specializing in extracting relevant information from provided documents. Your task is to identify and extract sentences from the documents as much as possible that are most directly useful for answering the given question. Rank the sentences in order of relevance, with the most relevant sentence listed first. Preface each sentence with its sequence number as follows:\\
Sentence 1: \\
......\\
Sentence n: \\

Question:\\
\verb|{Question}|\\
\\
Documents:\\
\verb|{Documents}|\\
\end{tcolorbox}

\label{tab:prompt1}
\end{table}

\begin{table}[H]
\caption{Prompt for the Adaptive Truncator.}
\begin{tcolorbox}
You are a highly skilled assistant specializing in optimizing language model efficiency by truncating context based on question complexity and document quality. Given a question and a ranked list of sentences, identify and retain the most relevant ones while truncating the irrelevant sentences.\\

Question:\\
\verb|{Question}| \\
\\
Ranked List:\\
\verb|{Ranked List}|\\
\end{tcolorbox}
\label{tab:prompt2}
\end{table}

\section{More Experimental Settings}

\subsection{Datasets}
\label{Appendix:dataset}
We conduct experiments on three widely used QA datasets that cover both single-hop and multi-hop question-answering scenarios. Table~\ref{qa-datasets} summarizes the key statistics of these datasets. Specifically, \textbf{NQ} (Natural Questions) and \textbf{TriviaQA} are representative single-hop datasets, where each question can typically be answered using information from a single passage retrieved from the corpus. These datasets primarily evaluate a model’s ability to locate and extract factual evidence efficiently. In contrast, \textbf{HotpotQA} is a challenging multi-hop dataset that requires integrating and reasoning over multiple pieces of evidence distributed across different documents to derive the final answer. This dataset is particularly useful for testing a model’s reasoning and compositional understanding capabilities. Together, these datasets provide a comprehensive benchmark for evaluating both the retrieval quality and reasoning robustness of our proposed method under diverse task settings.
\begin{table}[t]
\centering
\small
\renewcommand{\arraystretch}{1.1}
\caption{Statistics for the datasets.}
\begin{tabular}{lcccc}
\toprule[1.1pt]
\textbf{Dataset} & \textbf{Type}&\textbf{\# Train} & \textbf{\# Dev} & \textbf{\# Test} \\
\midrule
NQ      &  single-hop & 79.1k  & 8.7k  & 3.6k  \\
TriviaQA &  single-hop& 78.7k  & 11.3k  & 8.8k \\
HotpotQA  & multi-hop& 88.9k  & 5.6k  & 5.6k  \\
\bottomrule[1.1pt]
\end{tabular}
\label{qa-datasets}
\end{table}

\subsection{More Implementation Details }
\label{Appendix:settings}
We use LLaMA3-8B-Instruct~\footnote{\url{https://huggingface.co/meta-llama/Meta-Llama-3-8B-Instruct}}~\cite{llama3}, Qwen3-14B~\footnote{\url{https://huggingface.co/Qwen/Qwen3-14B}}~\cite{qwen3}, and LLaMA3.3-70B-Instruct~\footnote{\url{https://huggingface.co/meta-llama/Llama-3.3-70B-Instruct}}~\cite{llama3} as the generators, covering medium, large, and ultra-large LLMs. All of these models demonstrate strong performance across various tasks and exhibit high flexibility. To reduce computational cost and latency, our clue extractor and adaptive truncator are based on LLaMA3.2-3B-Instruct~\footnote{\url{https://huggingface.co/meta-llama/Llama-3.2-3B-Instruct}}~\cite{llama3}. We apply the LoRA method~\cite{hu2021lora} for fine-tuning, an efficient low-rank adaptation technique that reduces the computational cost of parameter updates while maintaining model performance. LoRA is applied to both the clue extractor and adaptive truncator. We train the models on two NVIDIA RTX PRO 6000 GPUs for 12 epochs. The initial learning rate is set to \(1\times10^{-4}\), and the batch size is 4. Gradient accumulation is employed to simulate larger effective batch sizes and improve training stability. The best model is selected based on validation set performance. During fine-tuning, the models are trained on the three QA datasets using a KNN-based sentence selection method. Data preprocessing is accelerated with 16 parallel workers per training epoch. The maximum input length is set to 4096 tokens to accommodate long-context information. 
For the clue reranker, we employ Sentence-BERT~\cite{reimers2020making} with the distilbert-base-uncased model to generate high-quality sentence embeddings for computing sentence similarity. Training uses the Adam optimizer with a batch size of 64, a learning rate of \(2\times10^{-5}\), and 1000 warm-up steps, and runs for 4 epochs.

\begin{figure}[!t]
\centering
  \includegraphics[width=1.0\columnwidth]{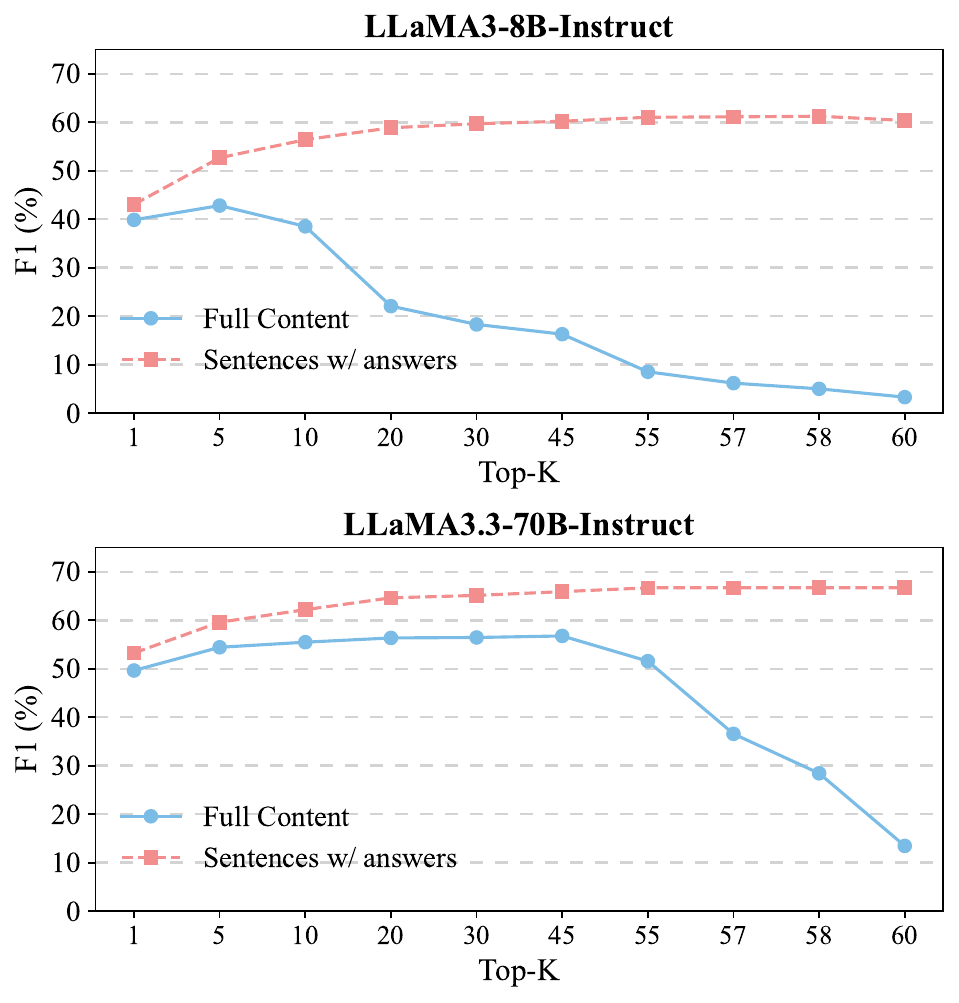}
\caption{F1 performance of LLaMA3-8B-Instruct (Top) and LLaMA3.3-70B-Instruct (Bottom) across varying Top-K retrieval settings. The retriever is Contriever-MS MARCO. Performance is shown for two strategies: Full Content and Sentences w/ answers.}
  \label{F1_comparison}
\end{figure}

\section{More Experimental Results on LLaMA3-70B}
\label{Appendix:70b}

To highlight the generalization capability of our method, we introduce a new retriever, \textbf{Contriever-MS MARCO}~\cite{msmarco,izacard2021unsupervised}, which retrieves the Top-60 documents per query for the experiments below.

\begin{table*}[!t]
\renewcommand{\arraystretch}{1.0}
  \centering
   \setlength\tabcolsep{14pt}
\renewcommand\arraystretch{1.05}

 \caption{
Experimental results (\%) on three benchmark datasets using LLaMA3.3-70B-Instruct as the generator. The retriever is DPR. The experimental setup follows Section~\ref{imple}. All baselines and our method are conducted using the same test sets and retrieval corpus. 
}

\resizebox{0.95\textwidth}{!}{
\scalebox{0.4}{
  \begin{tabular}{l ccccccccc}
     \toprule[1.1pt]
\multirow{2}{*}{\textbf{Method}} & \multicolumn{3}{c}{\textbf{NQ}} &\multicolumn{3}{c}{\textbf{TriviaQA}} & \multicolumn{3}{c}{\textbf{HotpotQA}}   \\
 \cmidrule(lr){2-4} \cmidrule(lr){5-7} \cmidrule(lr){8-10}
    &  \textbf{SubEM}  & \textbf{F1}  & \textbf{ CR}  & \textbf{SubEM} & \textbf{F1} &\textbf{CR}  & \textbf{SubEM}  & \textbf{F1} & \textbf{CR} \\

    \hline
       \rowcolor[HTML]{F0F0F0}
     \multicolumn{10}{c}{\textbf{\textsc{LLaMA3.3-70B-Instruct}}} \\
          
Naive Generation & 40.99 & 45.77 & -  & 73.36 & 76.74 & -  &28.63 & 35.55 & -  \\
  \cdashline{1-10}

  BM25  & 41.25 & 45.59 & 4.1$\times$ & 69.44  & 69.02 & 4.3$\times$ & 28.15 & 33.76 & 4.4$\times$ \\
  Bge-reranker  & 43.94 & 50.22 & 4.1$\times$ & 70.29  & 71.34 & 4.3$\times$ & 29.13 & 35.83 & 4.5$\times$  \\  

  RECOMP-extr &  43.60& 50.03 & 11.97$\times$ & 71.91 & 76.63 & 10.91$\times$ & 29.61 & 36.25 & 8.33$\times$  \\

    \cdashline{1-10}
  LongLLMLingua  & 48.39 & 53.67 & 4.56$\times$ & 74.37 & 78.12 & 4.18$\times$ & 31.03 & 38.42 & 4.45$\times$ \\
  
   Selective-Context & 50.14 & 55.79 & 2.6$\times$ & 74.21 & 77.03 & 2.7$\times$ & 30.88 & 37.93 & 2.7$\times$ \\
   EXIT & 47.25 & 51.83 & 14.16$\times$ & 66.06 & 67.43 & 12.78$\times$ & 27.48 & 32.25 & \textbf{15.43$\times$ } \\
  
    \cdashline{1-10}

 RECOMP-abs & 47.39 & 51.89 & 11.12$\times$ & 68.81 & 68.77 & 11.25$\times$ & 27.95 & 34.75 & 7.91$\times$ \\
 Refiner  & 49.22 & 54.02 & 10.97$\times$ & 72.85 & 77.29 & 12.63$\times$ & 30.42 & 37.12 & 7.65$\times$  \\
 BottleNeck & 49.53 & 54.44 & 14.32$\times$ & 72.96 & 77.46 & 13.17$\times$ & 30.75 & 37.76 & 13.21$\times$ \\

\cdashline{1-10}
 \textbf{Ours ($\epsilon = 0$)} & \textbf{50.35} & \textbf{56.21} & \textbf{17.19$\times$ }& \textbf{75.11} & \textbf{79.94} & \textbf{14.16$\times$} & \textbf{31.36} & \textbf{38.75} & 14.73$\times$ \\

      \bottomrule[1.1pt]
      \end{tabular} 
  }
    }
 
  \label{tab:70b}
\end{table*}
\subsection{Full Content vs. Sentences w/ answers}

Figure~\ref{F1_comparison} presents the F1 performance of LLaMA3-8B-Instruct and LLaMA3.3-70B-Instruct under different Top-K retrieval settings, where documents are retrieved using the \textbf{Contriever-MS MARCO} retriever. Two input strategies are compared: \textit{\textbf{Full Content}} and \textit{\textbf{Sentences w/ answers}}. For both model scales, performance initially improves as more documents are retrieved but later declines, with the larger model showing a delayed inflection point. Overall, the Sentences w/ answers strategy consistently surpasses the Full Content strategy across all Top-K configurations. Moreover, as the number of retrieved documents increases, the performance gain provided by Sentences w/ answers gradually diminishes and stabilizes, suggesting that this strategy offers robust and efficient context selection even under large-scale retrieval conditions.

\subsection{Comparison with Baselines}
Following the same experimental setting described in Section~\ref{imple}, we conduct experiments on NQ, TriviaQA, and HotpotQA. We use the \textbf{DPR} retriever to retrieve the Top-5 documents and adopt LLaMA3.3-70B-Instruct as the generator. All baselines and our method are conducted using the same test sets and retrieval corpus. The results are summarized in Table~\ref{tab:70b}. Across all three datasets, our method consistently outperforms various baseline approaches. This demonstrates that our approach effectively selects the most informative context from retrieved documents, enhancing performance on downstream generation tasks. Moreover, the consistent performance gains highlight the stability and strong generalization ability of our method across different task distributions, making it applicable in a wide range of retrieval and generation scenarios.

\subsection{Comparison with Full Content Strategy}
This section evaluates the generalization of our method across retrievers on the NQ dataset. We train with documents from the DPR retriever and test on the Contriever-MS MARCO retriever, simulating cross-retrieval shifts. Two extractors, LLaMA3.2-3B-Instruct (3B Extractor) and LLaMA3-8B-Instruct (8B Extractor), are compared against the Full Content baseline across Top-1 to Top-60 retrieved documents to analyze context-scale effects.

\begin{figure}[t]
\centering
  \includegraphics[width=0.85\columnwidth]{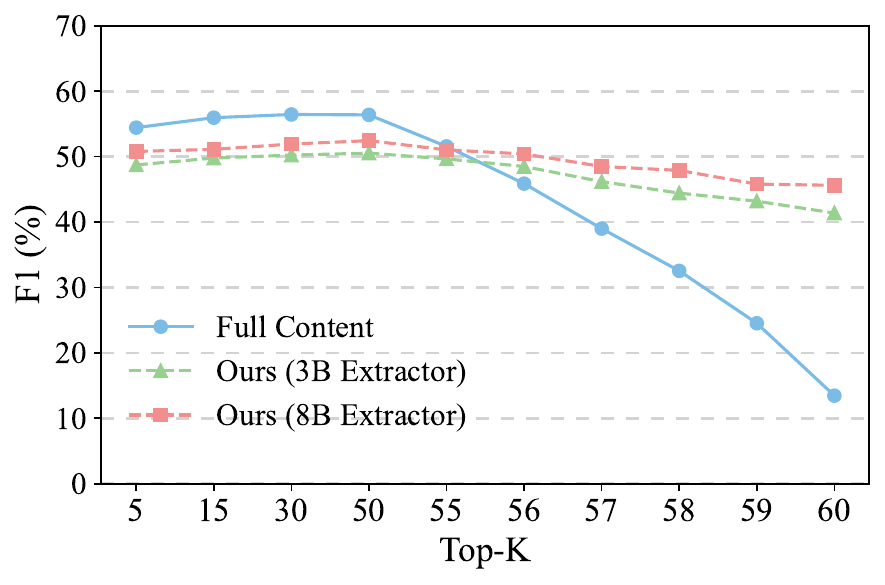}
\caption{F1 scores on the NQ dataset under different Top-K retrieval settings for three strategies: Full Content, 3B Extractor, and 8B Extractor. The retriever is Contriever-MS MARCO. We use LLaMA3.3-70B-Instruct as the generator and F1 for evaluation.}
  \label{top}
\end{figure}

As shown in Figure~\ref{top}, when the number of retrieved documents is small, our method performs slightly worse than the Full Content strategy. This is primarily because the LLaMA3.3-70B model itself has a sufficiently large context window and strong robustness, allowing it to effectively utilize information even under small-scale contexts. However, as the number of retrieved documents increases, the performance of the Full Content approach first rises and then declines, reflecting that the advantage of a large context window diminishes as document scale grows. In contrast, our method consistently outperforms the Full Content strategy when more documents are retrieved, indicating that selective truncation and compression can more effectively filter and leverage key information, thereby enhancing generation performance and stability under large-scale contexts. 
This pattern suggests that while smaller contexts may suffice for large LLMs, effective extraction of key sentences becomes increasingly important as more information is available. 

Increasing the extractor size from 3B to 8B improves performance but increases latency, highlighting a trade-off between efficiency and quality. Larger extractors better select key information for large-scale contexts, whereas smaller extractors may be preferable under limited resources or real-time constraints.

\begin{figure}[!t]
\resizebox{0.48\textwidth}{!}{\includegraphics{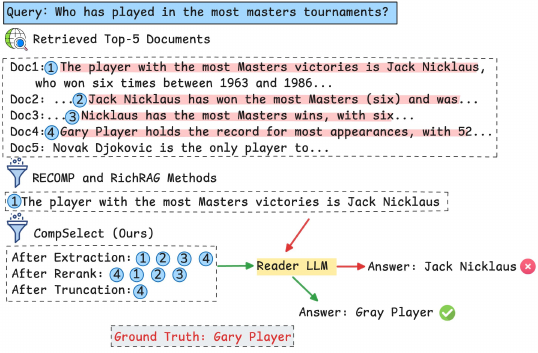}}
\centering
\caption{An illustration of the challenge in locating accurate answer clues. While baselines RECOMP and RichRAG select an incorrect clue from the first document, our method identifies the correct clue from the fourth via extraction, reranking, and truncation.
}
\label{fig:case_study}
\end{figure}
\section{Case Study}
As shown in Figure~\ref{fig:case_study}, our method highlights potential clues (red background), reranks them, and surfaces the correct answer clue at the top, while filtering out redundant ones to improve the information density of reasoning clues for RAG.

\end{document}